\documentclass[preprint,12pt,3p,times]{elsarticle}

\usepackage{amssymb}
\usepackage{multirow}
\usepackage{booktabs}
\usepackage{amsmath}
\usepackage{amsthm}
\usepackage{booktabs}
\usepackage{float}
\usepackage{setspace}
\usepackage[switch]{lineno}
\usepackage{xcolor}
\usepackage[procnumbered,linesnumbered,ruled]{algorithm2e}

\makeatletter
\def\ps@pprintTitle{
	\let\@oddhead\@empty
	\let\@evenhead\@empty
	\def\@oddfoot{Preprint.}
	\let\@evenfoot\@oddfoot
}
\makeatother

\begin{document}

\begin{frontmatter}

\title{Unlabeled Action Quality Assessment Based on Multi-dimensional Adaptive Constrained Dynamic Time Warping}

\author{Renguang Chen\fnref{label1}}
\author{Guolong Zheng\fnref{label2}}
\author{Xu Yang\fnref{label2}\corref{cor1}}
\author{Zhide Chen\fnref{label1}\corref{cor1}}
\author{Jiwu Shu\fnref{label2}\fnref{label3}}
\author{Wencheng Yang\fnref{label4}}
\author{Kexin Zhu\fnref{label5}}
\author{Chen Feng\fnref{label6}}

\address[label1]{College of Computer and Cyber Security, Fujian Normal University, Fuzhou, China}
\address[label2]{College of Computer and Data Science, Minjiang University, Fuzhou, China}
\address[label3]{Department of Computer Science and Technology, Tsinghua University, Beijing, China}
\address[label4]{School of Mathematics, Physics and Computing, University of Southern Queensland,Toowoomba, Australia}
\address[label5]{Department of Computer Science and Engineering, National Sun Yat-sen University, Taiwan, China}
\address[label6]{Department of information engineering,Fuzhou Polytechnic, Fuzhou, China\vspace{-2em}}

\cortext[cor1]{Corresponding author. Email: xu.yang@mju.edu.cn}

\begin{abstract}
    The growing popularity of online sports and exercise necessitates effective methods for evaluating the quality of online exercise executions. Previous action quality assessment methods, which relied on labeled scores from motion videos, exhibited slightly lower accuracy and discriminability. This limitation hindered their rapid application to newly added exercises. To address this problem, this paper presents an unlabeled Multi-Dimensional Exercise Distance Adaptive Constrained Dynamic Time Warping (MED-ACDTW) method for action quality assessment. Our approach uses an athletic version of DTW to compare features from template and test videos, eliminating the need for score labels during training. The result shows that utilizing both 2D and 3D spatial dimensions, along with multiple human body features, improves the accuracy by 2-3\% compared to using either 2D or 3D pose estimation alone. Additionally, employing MED for score calculation enhances the precision of frame distance matching, which significantly boosts overall discriminability. The adaptive constraint scheme enhances the discriminability of action quality assessment by approximately 30\%. Furthermore, to address the absence of a standardized perspective in sports class evaluations, we introduce a new dataset called BGym.
\end{abstract}


\end{frontmatter}

\section{Introduction}
\label{Introduction}
With the advancements in online sports instruction, the extensive volume of video homework submissions often overwhelms physical education instructors, inhibiting timely feedback and objective assessment. Automated action quality assessment scoring systems can address these issues by providing accurate and real-time evaluation.

One approach to action quality assessment is through action matching. A suitable method for action matching is Dynamic Time Warping (DTW). It provides an optimal monotonically aligned mapping between two input sequences through a dynamic programming process. In the context of the DTW, Euclidean distance is employed to compute the distance between two points in space. However, applying the DTW to scoring issues results in a significantly low level of discriminability. This is because the algorithm should ideally assign lower scores to actions that differ more significantly from the standard~\cite{Ying2021openpose}, such as comparing a model video of a jumping action to a student's video where the student remains standing still. In such cases, the score should ideally be below 60. Because different types of videos are videos with very different movements, they should be given a low score in the evaluation. But utilizing Euclidean distance with DTW merely distinguishes the action without assigning it an appropriately low score. 

To address similar challenges, Chang et al.~\cite{chang2021learning} introduced Discriminative Prototype DTW, which employs a discriminative loss to handle the temporal recognition of a sequence spanning multiple classes. Effenberg et al.~\cite{effenberg2016movement} calculated the distance value between the technology of the model and the technology of each parameter of the participant in the training process through the DTW algorithm. Then, according to the distance value of each parameter calculated by DTW algorithm, the binding force index (FI), the general distance value (GDV) is constructed according to the specific formula. To evaluate the effect of motor sonification on motor learning. Manousaki et al.~\cite{Manousaki2021} three DTW variants (OpenEnd-DTW, Soft Dynamic Time Warping, and Global Alignment Kernel) for time series alignment, Action labels are predicted by comparing the time series of incomplete actions with the time series performed by known prototype actions.

Another method for assessing action quality is regression-based approaches. Pirsiavash et al., \cite{pirsiavash2014assessing}, was one of the first to introduce deep learning to assess movement quality in sports, using regression models based on spatiotemporal pose features. Specifically, low level features and high level features based on human posture were used, quality assessment was modeled as a supervised regression problem, and linear support vector regression was used for training. Parmar et al.~\cite{parmar2019action} proposed learning a single model across multiple actions which utilizes the C3D model for feature extraction. Farabi et al.~\cite{farabi2022improving} employed the ResNet network for feature extraction and utilized the Weight-Decider technique to handle downstream scoring tasks, demonstrating its capability to handle 32-frame segments.
However, these end-to-end models~\cite{parmar2019action,farabi2022improving}, as the depth of the network increases, the performance of the network does not improve as expected, and the representation capability of features is gradually weakened or lost, resulting in performance degradation.

\begin{figure*}[t]
	\centering
	\includegraphics[width=400pt]{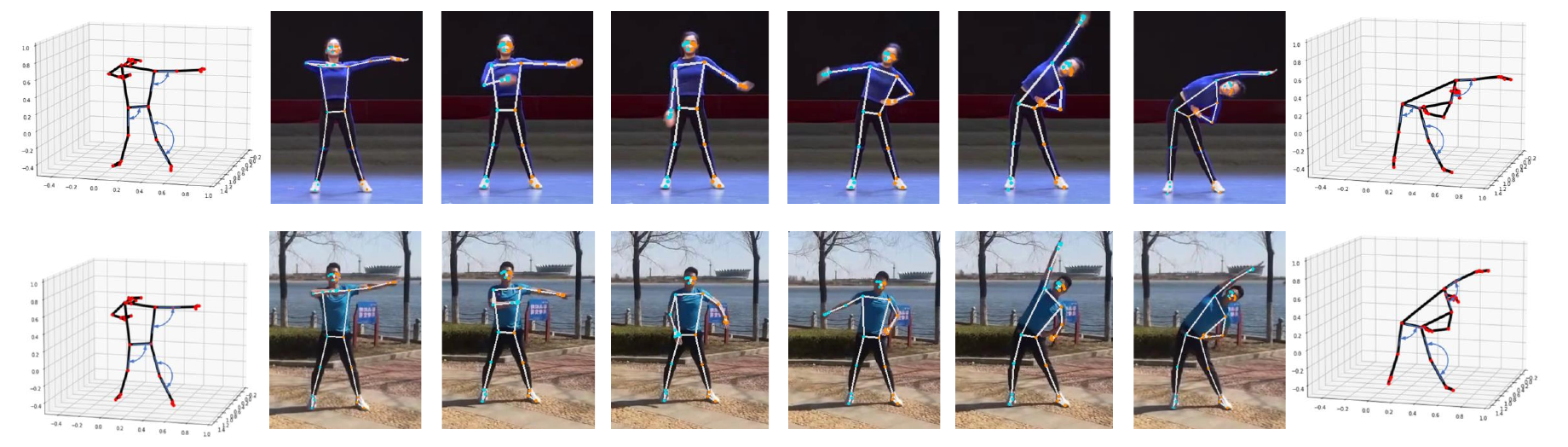}
	\caption{Action matching of two videos. The upper and lower parts respectively represent the extracted 2D and 3D joint points of the template video frame and the test video frame. Action matching is performed by comparing the differences between two videos.}
	\label{figure::matching}
\end{figure*}

Tang et al.~\cite{tang2020uncertainty} introduced the I3D-MLP scheme, which leverages the I3D model for extracting features. They further implemented a scoring distribution method to accurately represent the probability of action quality assessment scores, yielding precise outcomes. Building upon Tang et al.'s foundation, Zhang et al.~\cite{zhang2023auto} introduced the I3D-DAE scheme. This innovative approach includes a plug-and-play module called the Distribution Auto-Encoder (DAE), which enhances the accuracy of mapping videos to scores. Zhou et al.~\cite{zhou2023hierarchical} introduced a sophisticated hierarchical graph convolutional network (HGCN), designed to address the challenges that arise when video sequences are divided into uniformly sized clips. This division often results in confusion within clips and incoherence between them. Their approach significantly improves the discriminative features of the videos, thereby elevating the performance of action quality assessment. However, it should be acknowledged that these supervised approaches require a significant amount of work to construct labels for action category datasets, often relying on expert judges to assign scores to corresponding actions.

To overcome these issues, this paper introduces a novel action quality assessment algorithm MED-ACDTW. Firstly, in response to the reliance on labeled regression schemes, this work proposes a comparative approach that evaluates the representation of test videos by comparing them to template videos. This indirect evaluation of action quality reduces the dependence on labels and allows for the utilization of template videos as an alternative label attribute, offering greater flexibility for real-world applications. In practical situations, acquiring a large amount of data with accurate labels is challenging, but action quality assessment can be achieved by collecting a set of standard videos as references, as depicted in Figure \ref{figure::matching}.

Secondly, to address the issue of inadequate accuracy in DTW-based evaluation of time series, this paper proposes two solutions for feature and frame distance calculation. Regarding features, the focus is on joint angle features, directional features, and upper-lower limb coordination features. This paper introduces directional features and upper-lower limb coordination features for the hip joint for the first time. Moreover, relying solely on 2D or 3D features has limitations, such as inaccurate recognition due to occlusion in 2D or insufficient depth information in 3D. Therefore, this paper combines features from both modalities, extracting joint angle features and directional features (excluding hip joint angles) from 2D skeletal joints, and capturing limb correlations and additional local features, including hip joint and shoulder joint rotations, from 3D skeletal joints. To handle the distance calculation issue arising from these features, the paper proposes the MED module, which employs a ratio-based approach to measure the impact of each feature.

Thirdly, classic DTW algorithms are not robust to outliers, as the ideal DTW path should follow a diagonal pattern, while outliers can distort the path. This distortion leads to excessive stretching and compression~\cite{wan2017adaptive}. To tackle this issue, adaptive constraints are applied in the context of action quality assessment. This is the first application of adaptive constraints in motion sequences, and we have adjusted and validated it for this specific scenario. Specifically, we make the following contributions:
\begin{enumerate}
	\item[(1)] 
	We design the MED-ACDTW algorithm, which achieves unlabeled action matching by comparing template videos and test videos. Eliminating the need for laborious annotation operations such as scoring difficulty levels in action videos.
	\item[(2)] 
	By applying adaptive constraints to the action sequence, we refine the accuracy of the DTW algorithm in measuring the distance between trajectories, effectively addressing the challenge of multiple frames mapping to a single frame within an action time series.
	\item[(3)] 
	We construct new action matching features based on skeletal points from multiple dimensions and angles. These features enable the fine-grained representation of human body pose features. 
	To apply the results of features to DTW, we propose the MED method, which offers greater discriminative power than the Euclidean distance.
	\item[(4)] 
	We create a new dataset for action quality assessment, consisting of 153 cropped videos, each lasting 3 seconds, along with corresponding evaluation labels. This dataset addresses the lack of standardization in the field of physical education examinations.
\end{enumerate}

\section{Related Work}\label{Sec::RW}

\subsection{Skeleton Data Acquisition}
In the field of skeleton information acquisition, researchers have explored multiple modalities. First of all, MediaPipe ~\cite{lugaresiMediaPipeFrameworkPerceiving}, a lightweight pose estimation model, has demonstrated high accuracy and low latency pose detection, making it well-suited for mobile devices. Zhang et al.~\cite{zhangMediaPipeHandsOndevice2020} utilized MediaPipe in a framework for developing cross-platform machine learning solutions. Their proposed model and pipeline architecture exhibited real-time inference speeds on mobile GPUs and high prediction quality. MediaPipe also finds applications in medical analysis. For instance, Güney et al.~\cite{guneyVideoBasedHandMovement2022} employed MediaPipe to quantitatively evaluate hand movements in patients with Parkinson's disease, proposing automatic tremor analysis.

\subsection{Feature Information Representation}

After processing the skeleton data, the derived features can capture various postures and reflect the nuances of human movement. Hu et al.~\cite{hu2015jointly} constructed pose features based on the relative positions of skeletons, considering the spatiotemporal trajectory of their movements. Rahmami~\cite{rahmani2014real} utilized the displacement information of skeletons to compute the maximum sweep area, which represents the postural characteristics of the human body. The directional features of joint limbs can encode joint angle information. Gu et al.~\cite{gu2012human} calculated the direction of the displacement vector using 15 skeletons, extracting joint angle features that represent the human torso and reflect the relative direction of the joints. Bakhat et al.~\cite{bakhat2023katz} observed the joint order of 3D joints, extracted temporal information, and utilized zero-crossing features between current and previous frames to derive reliable features. They proposed an OMKZ scheme using ordered, modified Katz centrality and zero-cross information for human motion recognition.

	\subsection{DTW Algorithm}\

DTW has been widely used as a distance measure for time series in various fields, such as clustering, classification, and similarity search \cite{ding2008efficient,papapetrou2011embedding}. Researchers have utilized DTW to compare action sequences \cite{effenberg2016movement} and used the comparison results as a score for action evaluation, enabling fast similarity searches \cite{keogh2004indexing,kruger2010fast}.  Zhao et al.~\cite{zhao2018shapedtw} proposed shapeDTW, which augments DTW by considering point-wise local structural information. It was implemented as a distance measure in the nearest neighbor classifier scheme and demonstrated superior performance compared to DTW. To address the issues of over-stretching and over-compression, Niennattrakul et al.~\cite{niennattrakul2007clustering} introduced an adaptive window restriction method to determine the optimal warping path with limited stretching and compression. However, this approach does not guarantee the continuity of the optimal path. Another notable contribution is the Adaptive Constrained DTW algorithm~\cite{li2020adaptively}, which focuses on accurately calculating the distance between trajectories.

\begin{figure*}[htbp]
	\centering
	\includegraphics[width=\textwidth]{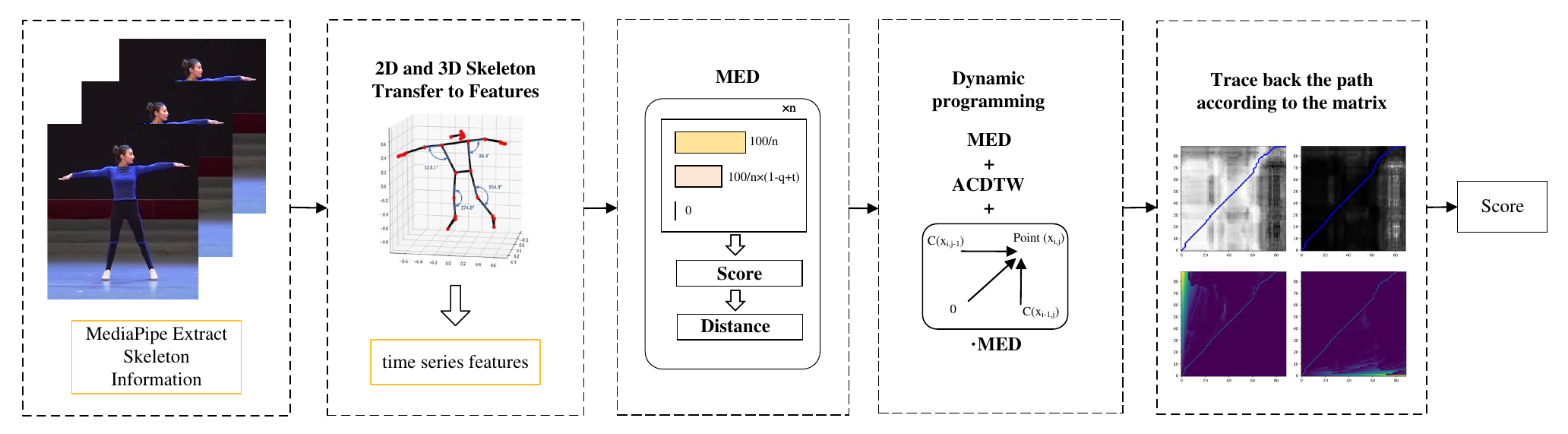}
	\caption{The overall structure and workflow of the MED-ACDTW, Module 1 illustrates the process of extracting continuous frames of 2D and 3D keypoints using MediaPipe. The extracted keypoints then enter the Human Feature Construction module, which includes four types of features. The resulting frame-based time series features are input into the third module for score and distance calculation. The computed distances are then fed into the ACDTW module with adaptive constraints to obtain the constraint matrix and score matrix. Backtracking is performed to calculate the overall score based on this matrix. }
	\label{figure::approach}
\end{figure*}

\section{Our Approach}\label{Sec::approach}

In this section, we first introduce how to extract the features of skeleton joint points. Then, we describe the MED-ACDTW method in detail. The architecture of MED-ACDTW for action quality assessment is shown in Figure \ref{figure::approach}.

\subsection{Feature Extraction}
When relying solely on skeleton point coordinates for DTW, the accuracy can be insufficient due to individual variations in body structure and movements. To overcome this limitation, we propose constructing both local features, which capture the relationships between different body joints, and global features, which reflect overall movement coordination. These features are designed to accommodate individual variations in body structure and movements. 
These features are designed to accommodate individual variations in body structure and movements. The detailed construction of these features is outlined in Table \ref{table::feature_detail}. Barycentric and Upper and Lower Limb Linkage connect the global limbs, while Joint and Direction tend to connect the local limbs.

\subsubsection{Joint angle feature}
\begin{figure}[ht]
	\centering
	\includegraphics[width=400pt]{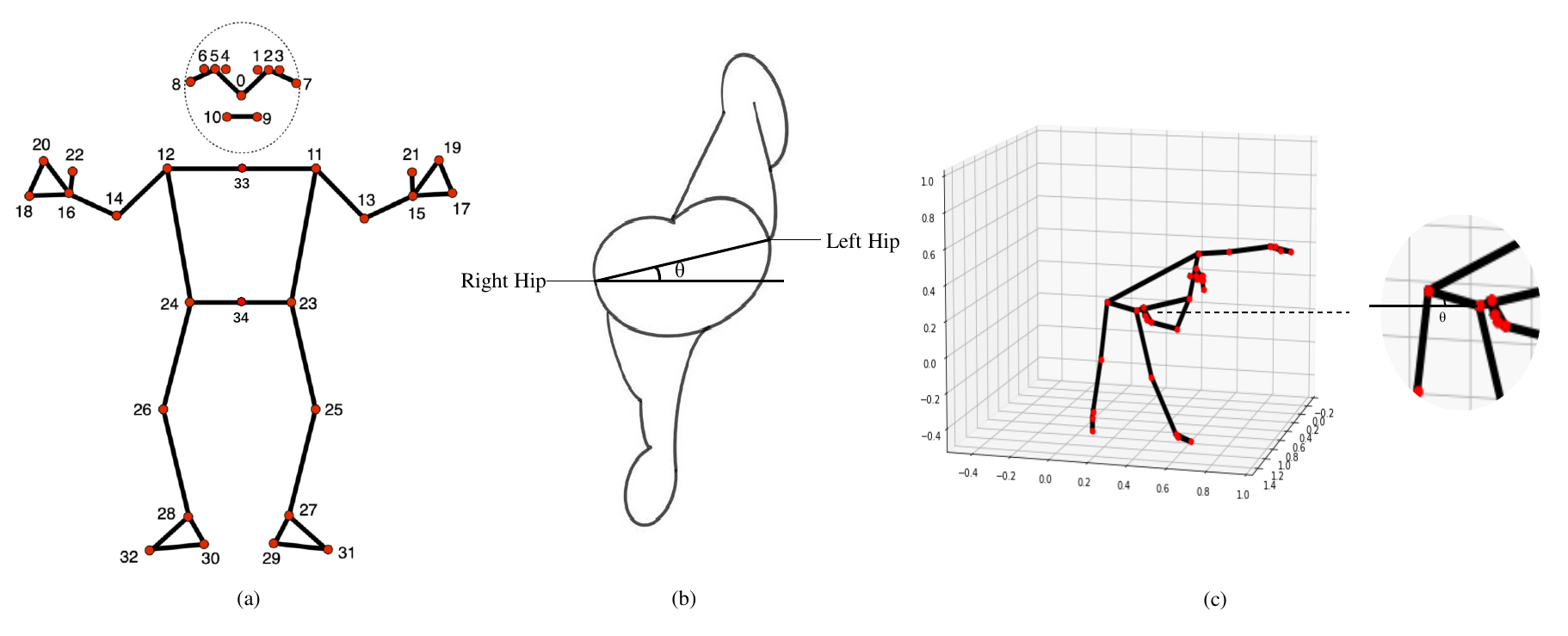}
	\caption{Feature presentation. (a) represents the joint points of humans identified by MediaPipe. (b) represents the Pelvic Horizontal Angle. (c) represents the Pelvic Rotation Angle.
	}
	\label{figure::feature1}
\end{figure}

For feature construction, joint points are shown in Figure \ref{figure::feature1}(a), where nodes 0 to 32 represent default nodes in MediaPipe. In the angular feature construction, we build two features. First, we aim to establish connections between adjacent joints in the human body. To achieve this, we calculate the angles between 13 adjacent joints, focusing on key areas such as elbows, shoulders, and knees. The second involves characterizing the deviation of limb block center points. We consider each of the four limbs as a separate region, depicted by a polygon formed by specific joint points within that region. We then compute the center point of each region and calculate the Angle between that point and the spine vector. The calculation method is shown in Eq.\eqref{eq5} and Eq.\eqref{eq6}, where $p_0$ represents the coordinates of the center point. $x_i, y_i, z_i$ respectively represent the vertices of the polygon. $\overrightarrow{M}$ stands for spine vector.

\begin{equation} \label{eq5}
	p_0=\left(\frac{x_1+x_2+x_3+x_4}{4}, \frac{y_1+y_2+y_3+y_4}{4}, \frac{z_1+z_2+z_3+z_4}{4}\right)
\end{equation}

\begin{equation} \label{eq6}
	\theta=\arccos \left(\frac{\overrightarrow{M} \cdot \overrightarrow{op_0}}{|\overrightarrow{M}| \times|\overrightarrow{op_0}|}\right)
\end{equation}

\subsubsection{Barycentric feature}

In order to determine the barycentric feature, we employed the coefficient method~\cite{Ying2021openpose}. The process implies multiplying the coordinates of each joint point by the corresponding inertia parameters of the human body to yield the body's center of gravity or barycenter. Also, calculate the angle between the Barycentric and the spine to form a feature.

\begin{table}[!htbp]
	\begin{center}
	\caption{All feature details}
	\label{table::feature_detail}
	\begin{tabular}{l|l|l}
		\toprule
		Type                                                                                     & Name                                                                                                 & Num \\ 
		Barycentric                                                                              & Human Body Angle                                                                                     & 1   \\ \midrule
		\multirow{2}{*}{Joint Angle}                                                             & Limb Mass Center Point                                                                               & 4   \\ \cline{2-3} 
		& Limb Angle                                                                                           & 13  \\ \midrule
		\multirow{3}{*}{Direction}                                                               & Human Body 2D Direction                                                                              & 2   \\ \cline{2-3} 
		& \begin{tabular}[c]{@{}l@{}}Pelvic Horizontal Angle,\\ Pelvic Rotation Angle\end{tabular}     & 2   \\ \cline{2-3} 
		& \begin{tabular}[c]{@{}l@{}}Shoulder Horiz. Angle,\\ Shoulder Rotation Angle\end{tabular} & 2   \\ \midrule
		\multirow{4}{*}{\begin{tabular}[c]{@{}l@{}}Upper and Lower \\ Limb Linkage\end{tabular}} & Forearm \& Calf Angle                                                                        & 2   \\ \cline{2-3} 
		& Forearm \& Thigh Angle                                                                       & 2   \\ \cline{2-3} 
		& Foot L. \& Shoulder W. Ratio                                                                         & 1   \\ \cline{2-3} 
		& Wrist L. \& Shoulder W. Ratio                                                                        & 1   \\  \bottomrule
	\end{tabular}
\end{center}
\end{table}

\subsubsection{Direction feature}
The direction features are constructed using the angles of the hip and shoulders, as shown in Figure \ref{figure::feature1}(b) and (c). To construct the features for the pelvic horizontal angle and the pelvic anterior tilt angle, we calculate the angle between the offset direction of these body parts and the coordinate axes. The same methodology is applied to the shoulder angles, including the shoulder horizontal and anterior tilt angles. In the 2D context, we primarily use the hip and shoulder angles to remap the Boolean features relative to their angles to the coordinate axes.

Table \ref{table::feature_detail} presents a comprehensive breakdown of the feature details, consisting of a total of 30 distinct features. We compute both the central and angular features simultaneously in both 2D and 3D dimensions. As a result, the total count of features is 52, comprising 30 features in 2D and 22 features in 3D.

\subsubsection{Upper and lower limb linkage feature}
While the previously mentioned features mainly focus on the upper or lower limbs individually, there is a need for features that effectively bridge the upper and lower limbs.
To address this, we introduce new features based on the angles between the forearm and calf, and the forearm and thigh. These additional features enhance our ability to evaluate the linkage between the upper and lower limbs during actions such as standing upright or sitting. For the features that involve the forearm and calf, using the left side of the body as an example, $\overrightarrow{A B}$ represents the left forearm, $\overrightarrow{C D}$ represents the left calf. The linkage features of the left half of the body can be calculated using Eq.\eqref{eq7_1}.	Similarly, calculations are performed for the calf and thigh.

\begin{equation} \label{eq7_1}
	\theta=\arccos \left(\frac{\overrightarrow{A B} \cdot \overrightarrow{C D}}{|\overrightarrow{A B}| \times|\overrightarrow{C D}|}\right)
\end{equation}

Additionally, we utilize the ratio of the tensioned feet to the shoulder width and the ratio of the tensioned hands to the shoulder width to assess the tension in the limbs. These features are effective in distinguishing between upright and stretching motions. To appropriately integrate these features into the human posture features framework of this study, it is necessary to convert the non-radian numbers into radians. Consequently, we convert the ratios of the tension degree of feet to shoulders and hands to shoulders into radians. By performing these steps, we obtain features that are well-suited for posture matching.
The specific steps are as follows:

\begin{enumerate}
	\item[(1)] 
	Calculate the 3D distance between both ankle joints and the 3D distance between shoulder joints. 
	If the distance between the ankles is greater than half the distance between the shoulders, then $\theta_1=3 \pi / 2$; if it is less, $\theta_1=\pi / 2$. This provides the feature $\theta_1$ - the ratio of the feet's tension degree to the shoulder width.
	\item[(2)] 
	Calculate the 3D distance between the wrist and shoulder joints. 
	If the distance between the wrists is greater than $3/2$ the shoulder distance, then $\theta_2=3 \pi / 2$. 
	If it is less, $\theta_2=\pi / 2$. 
	This provides the feature $\theta_2$ - the ratio of the hand's tension degree to the shoulder width.
\end{enumerate}

\subsection{Action matching algorithm}

The Action matching algorithm incorporates the MED module to transform multi-dimensional features into feature score matrices and calculate inter-frame motion distances. This module is used to standardize the distance calculation, ensuring fair and consistent comparisons across all features, considering the significant variations in feature magnitudes and resulting disparate scoring ratios.

Furthermore, a dynamic programming matrix is generated using the concept of DTW, with the addition of adaptive constraints to penalize deviations from the path. A penalty is not added for direct diagonal movement, but moving upward or to the right incurs a penalty, thus making the approach adaptable to the motion time series.

Finally, the optimal solution is found using the backtracking method based on the distance matrix of the dynamic programming matrix, which is then mapped to the score matrix. The final score is obtained by accumulating the scores and computing the average.
	
\subsubsection{MED module}
\label{MED_Module}
The MED module is utilized to calculate the score between two frames. The following definitions are used:

\begin{itemize}
	\item 
	$\alpha$ and $\beta$ correspond to the number of frames in the template and test videos, respectively.
	
	\item
	$n$ represents the number of the template and test video frames.
	
	\item 
	$X=\left\{x_1, x_2, \cdots, x_n\right\}$ and $Y=\left\{y_1, y_2, \cdots, y_n\right\}$ represent the values of each feature in a test video frame and a template video frame, respectively.
	
	\item
	$l$ is the length of the scoring path in the 2D matrix. 
	
	\item
	$Score_{1,1}$, $Score_{1,2}, \cdots$, $Score_{\alpha, \beta}$ are the scores calculated for two frames. 
	Each pair of frames represents a single point on the two-dimensional matrix.
	
	\item
	$FS$ denotes the final score measured for the two videos. 
	
\end{itemize} 

To calculate the score, each feature score is computed individually.
As shown in Eq.\eqref{eq8}, the sum of all feature scores totals 100 points.
Next, the corresponding limb features of the template and the test video are subtracted to determine their difference. The ratio of this difference to the feature of the template video is then calculated. Based on the value of this ratio, denoted as $q$, the score for each feature is determined. According to Eq.\eqref{eq9} and Eq.\eqref{eq10}, if $q$ is less than the threshold parameter $t$, it is considered an acceptable error, and the score remains unchanged.

\begin{equation} \label{eq8}
	\begin{aligned}
		S_1=S_2=\ldots=S_n=\frac{100}{n}
	\end{aligned}
\end{equation}

\begin{equation} \label{eq9}
	\begin{aligned}
		q=\frac{\left|x_i-y_i\right|}{x_i}
	\end{aligned}
\end{equation}

\begin{equation} \label{eq10}
	\begin{aligned}
		S_i=\left\{\begin{array}{cc}
			S_i , & q \leq t \\
			S_i \times(1-q+t), & t<q \leq 1 \\
			0 , & q>1
		\end{array}\right.
	\end{aligned}
\end{equation}

The score for all features of two frames is calculated according to Eq.\eqref{equ:eq12}. The scores of all features are then summed to obtain the score between the two frames. Based on this score and Eq.\eqref{equ:eq13}, the distance between the two frames can be calculated, which replaces the Euclidean distance.

\begin{equation} \label{equ:eq12}
	Score_{i,j}=\sum_{k=1}^{n}S_{k}
\end{equation}

\begin{equation} \label{equ:eq13}
	MED(i,j)=\frac{1}{Score_{i,j}}\times100-1
\end{equation}

The score corresponding to each frame on the path is calculated using the ACDTW algorithm and the MED module. The sum of the scores on the path is then averaged to obtain the final score between the test video and the template video. This represents the score on the score path. The calculation of the path score is shown in Eq.\eqref{equ:eq14}. Algorithm \ref{alg:MED} for calculating feature distance instead of Euclidean distance is shown below.

\begin{equation} \label{equ:eq14}
	FS=\frac{\sum_{i=0,j=0}^{n,n}Score_{i,j}}{l}
\end{equation}

	\begin{algorithm}[htbp]
	\caption{Calculation of inter-frame distance via MED module}
	\label{alg:MED}
	\SetAlgoLined
	
	\KwIn{Template video feature $x_i(i=1\dots n)$, Test video feature $y_i(i=1\dots n)$}
	
	\KwOut{$med$} 
	
	$score=0$, $n=len(x_i)$
	
	\For{$i=1: n$}{
		$t= $parameters, ideally 0.2 \\
		$q=abs(x_i-y_i)/x_i$  \\
		\If{$q \leq t$}
		{
			$point\_score=100/n$
		}
		\ElseIf {$q > 1$}
		{
			$point\_score=0$
		}
		\Else
		{
			$point\_score=100/n \times (1 - q + t)$
		}
		$score=score+point\_score$
	}
	$med = score/100-1$ \\
	\Return $med$
\end{algorithm}

\subsubsection{Adaptive Constrained DTW}
For convenience, this paper initially provides the formula for the unrefined DTW, as shown in Eq.\eqref{equ:dtw}, to enable a comparison with the enhanced formula within the ACDTW. $ED$ represents the Euclidean distance between features between two frames.

\begin{equation} \label{equ:dtw}
	\begin{aligned}
		& DTW(1,1)=ED(1,1), \\
		& DTW(i,j)=ED(i,j)+\min\left\{\begin{matrix}DTW(i-1,j-1)\\DTW(i-1,j)\\DTW(i,j-1)\end{matrix}\right.
	\end{aligned}
\end{equation}

Transitioning from the foundational DTW approach, the ACDTW method introduces a novel adaptive penalty function designed to address the specific challenge of multiple frames mapping to a single frame within an action time series. This adaptive approach seeks to optimize the usage of each frame in the action sequence by expanding the penalty function as the number of uses increases~\cite{li2020adaptively}. We employ a adaptive penalty function that takes into account varying trajectory lengths and strengthens the penalty function via dual constraints. Eq.\eqref{equ:penalty} shows the adaptive penalty function.

\begin{equation} \label{equ:penalty}
	C(x_{i,j})=\frac{2\max(a,b)}{(a+b)}•N(x_{i,j})
\end{equation}

In this penalty function, $a$ and $b$ represent the lengths of the two action time series respectively. $N(x_{i,j})$ denotes the frequency of each point's participation in the matching process. Meanwhile, the penalty exhibits proportional growth corresponding to an increasing magnitude of difference between $a$ and $b$. This is because when two video frames significantly differ in length, they are likely not from the same category of video. As such, a larger penalty is applied to distinguish them. Thus, the value of the penalty function is directly proportional to the distance between the two time series, indicating a decreasing tolerance for many-to-one and one-to-many matches as the divergence between the two time series increases.

When the coefficient is zero, it corresponds to the standard DTW algorithm. Moreover, when the constraint coefficient is constant, $C(x_{i,j})$ is directly proportional to $N(x_{i,j})$. The computation of $N(x_{i,j})$ relies on two matrices, $T_{P}=\left(P_{i,j}\right)m\times n$ and $T_{Q}=\left(Q_{i,j}\right)m\times n$, which denote the frequency of participation in the matching process for $P$ and $Q$ respectively. The action sequence for frame $p_i$ participates $P_{i,j}$ times, and the action sequence for frame $q_i$ participates $Q_{i,j}$ times. The sum of these two participation counts yields the value of $N(x_{i,j})$. To prevent the occurrence of multiple frames corresponding to a single frame or a single frame corresponding to multiple frames. 

The state transition equation of the ACDTW is expressed in Eq.\eqref{equ:eq16}.	In this penalty function, $a$ and $b$ represent the lengths of the two action time series respectively. $N(x_{i,j})$ denotes the frequency of each point's participation in the matching process. Meanwhile, the penalty exhibits proportional growth corresponding to an increasing magnitude of difference between $a$ and $b$. This is because when two video frames significantly differ in length, they are likely not from the same category of video. As such, a larger penalty is applied to distinguish them. Thus, the value of the penalty function is directly proportional to the distance between the two time series, indicating a decreasing tolerance for many-to-one and one-to-many matches as the divergence between the two time series increases.

\begin{equation} \label{equ:eq16}
	\begin{aligned}
		&ACDTW(1,1)=MED(1,1), \\
		&ACDTW(i,j)=MED(i,j)+&	   \\
		&\min\left\{\begin{matrix}{ACDTW(i-1,j-1)}\\{ACDTW(i-1,j)+C(x_{i-1,j}) MED(i,j)}\\{ACDTW(i,j-1)+C(x_{i,j-1}) MED(i,j)}\end{matrix}\right.
	\end{aligned}
\end{equation}

If the trajectory from $(i-1,j-1)$ to $(i,j)$ visits unique indices, the penalty function is not employed in this step. If the path from $(i-1,j)$ to $(i,j)$ indicates the $q_j$ frame is reused, the MED incorporates a penalty function, setting $P_{i,j}=1$ and $Q_{i,j}=1+Q_{i-1,j}$. If the path from $(i,j-1)$ to $(i,j)$ suggests that the $p_i$ frame is reused, the MED similarly adds a penalty function, setting $Q_{i,j}=1$ and $P_{i,j}=1+P_{i-1,j}$. This approach allows the MED-ACDTW algorithm to handle multi-dimensional action time series while preventing multiple frames from corresponding to a single frame. The pseudo-code summary of the ACDTW algorithm is shown in Algorithm 2.

\begin{algorithm}[ht]
	\caption{ACDTW algorithm}
	\label{alg:ACDTW}
	\SetAlgoLined
	
	\KwIn{Template video feature $x_i(i=1\dots n)$, Test video feature $y_i(i=1\dots n)$}
	
	\KwOut{Optimal warping path, distance between $x$ and $y$} 
	
	Initialize: $ P_{1,1}=1, P_{i,j}=0,  i=2,\cdots,m, j=2,\cdots,n.$ 
	$ Q_{1,1}=1, Q_{i,j}=0,  i=2,\cdots,m, j=2,\cdots,n.$ 
	$ACDTW(1,1)=MED(1,1)$
	\;
	
	\For{$i=1: m$}{
		\For{$j=1: n$}{
			$D1=MED(i,j)+ACDTW(i-1,j-1)$ \\
			$D2=MED(i,j)+c(x_{i-1,j}) MED(i,j)+ACDTW(i-1,j)$\\
			$D3=MED(i,j)+c(x_{i,j-1}) MED(i,j)+ACDTW(i,j-1)$\\
			$ACDTW(i,j)=\min(D1,D2,D3)$\\
			
			$Optimal \  path[i,j] = min\_index((i-1, j), (i-1, j-1), (i, j-1))$ \\
			\If{$ACDTW(i,j)==D1$}
			{
				$P_{i,j}=1,Q_{i,j}=1$\;
			}
			\ElseIf {$ACDTW(i,j)==D2$}
			{
				$P_{i,j}=1,Q_{i,j}=1+Q_{i-1,j}$\;
			}
			\Else
			{
				$P_{i,j}=1+P_{i,j-1},Q_{i,j}=1$\;
			}
		}
	}
	\Return $path$, $P,Q$;
\end{algorithm}

\section{EXPERIMENTS}\label{Sec::Experiments}

In this section, we first introduce the metrics and datasets used in this article. Next, we apply the MED-ACDTW algorithm to the constructed dataset, optimizing parameters through ablation experiments. Then, we compare the performance of our algorithm against other mainstream methods. Finally, we present and interpret the results through visualization.

\subsection{Experimental setup}

The experimental setup involves matching an action with the highest score and comparing it against the true action for consistency. A match is successful if the same type of action achieves the highest score, implying that all scores of different action types are lower. Conversely, if a different action type achieves the highest score, it is considered a mismatch. Consequently, the matching success rate (accuracy) serves as an evaluation metric for our approach, as shown in Eq.\eqref{equ:eq17}.

Furthermore, to assess action quality, similar actions aim to score above 80, while different types of actions should score below 80. This distinction aids in applying the methods proposed in this study to practical scenarios. Therefore, we introduce an evaluation metric, $Rate80$, to indicate this discrimination effect. Eq.\eqref{equ:eq18} represents the total number of similar actions assigned a score above 80 and different types of actions assigned a score below 80.

In addition, to compare with some of the currently popular approaches, the expertise of professional sports instructors labels the dataset. This enables supervised training and inference of scores using the proposed method. After labeling the dataset, the performance of different approaches on the smaller dataset, BGym, is evaluated. The evaluation metric used is Spearman correlation, commonly employed in popular approaches. As shown in Eq.\eqref{equ:rho}. $\rho$ is the Spearman’s rank correlation coefficient. $d_i$ is the difference between the ranks of corresponding values from the two samples. $n$ is the number of samples.

\begin{equation} \label{equ:eq17}
	Acc=\frac{\text{Successful matching actions}}{\text{Total number of actions}}
\end{equation}

\begin{equation} \label{equ:eq18}
	Rate80=\frac{M+N}{\text{Number of action matches}}
\end{equation}

\begin{equation} \label{equ:rho}
	\rho = 1 - \frac{6 \sum d_i^2}{n(n^2 - 1)}
\end{equation}

\subsection{Dataset}\label{Sec::dataset}
The AQA dataset~\cite{parmar2019and} comprises sports videos from the Olympic Games, annotated with difficulty and scoring labels. However, the AQA dataset often suffers from frequent camera switches, resulting in inconsistent perspectives. Therefore, it is not suitable for addressing action matching problems in sports curriculum. To address this limitation, we propose the BGym dataset.

The BGym dataset is collected from the broadcast gymnastics course. Once this paper is accepted, the dataset will be made publicly available. The broadcast gymnastics routines were divided into eight sections: stretching exercises, chest expansion exercises, kicking exercises, lateral exercises, body rotation exercises, abdominal and back exercises, jumping exercises, and ending exercises. However, the final section is excluded, as some videos do not include it, leaving the template video with seven sections. 

We selected one of the most standardized videos to serve as a template. The first row of Table \ref{table::dataset} shows that the template video contains 7 actions, and the bottom row represents the 7 categories corresponding to the test video. Each exercise contains four eight-beat movements. We identified and collected the most representative movements from these eight-beat sequences, each lasting approximately 3 seconds. For instance, in the stretching exercise section, the selected three-second-long case includes flat-side stretching with hands, crouching with fists, hands-up stretching, and standing at attention. In addition, some beats in the four eight-beat movements are not executed with strict adherence to the standard. That is, certain beats exhibit significant deviations from the others. Consequently, three such instances are extracted from each practice session to serve as three videos in the dataset.

\begin{table}[H]\setlength{\tabcolsep}{1.6mm}
	\caption{Dataset overview}
	\label{table::dataset}
	\centering
	\begin{tabular}{ccc}
		\toprule
		
		\multicolumn{1}{l}{Number} & Video category               & \multicolumn{1}{l}{Amount} \\ \midrule
		-                          & Template Video               & 7                          \\ \midrule
		1                          & Lateral Exercises            & 24                         \\
		2                          & Chest Expansion Exercises    & 21                         \\
		3                          & Kicking Exercises            & 21                         \\
		4                          & Stretching Exercises         & 24                         \\
		5                          & Body Rotation Exercises      & 21                         \\
		6                          & Abdominal and Back Exercises & 21                         \\
		7                          & Jumping Exercises            & 21                         \\ \bottomrule
	\end{tabular}
\end{table}

\begin{figure*}[!htbp]
	\centering
	\includegraphics[width=480pt]{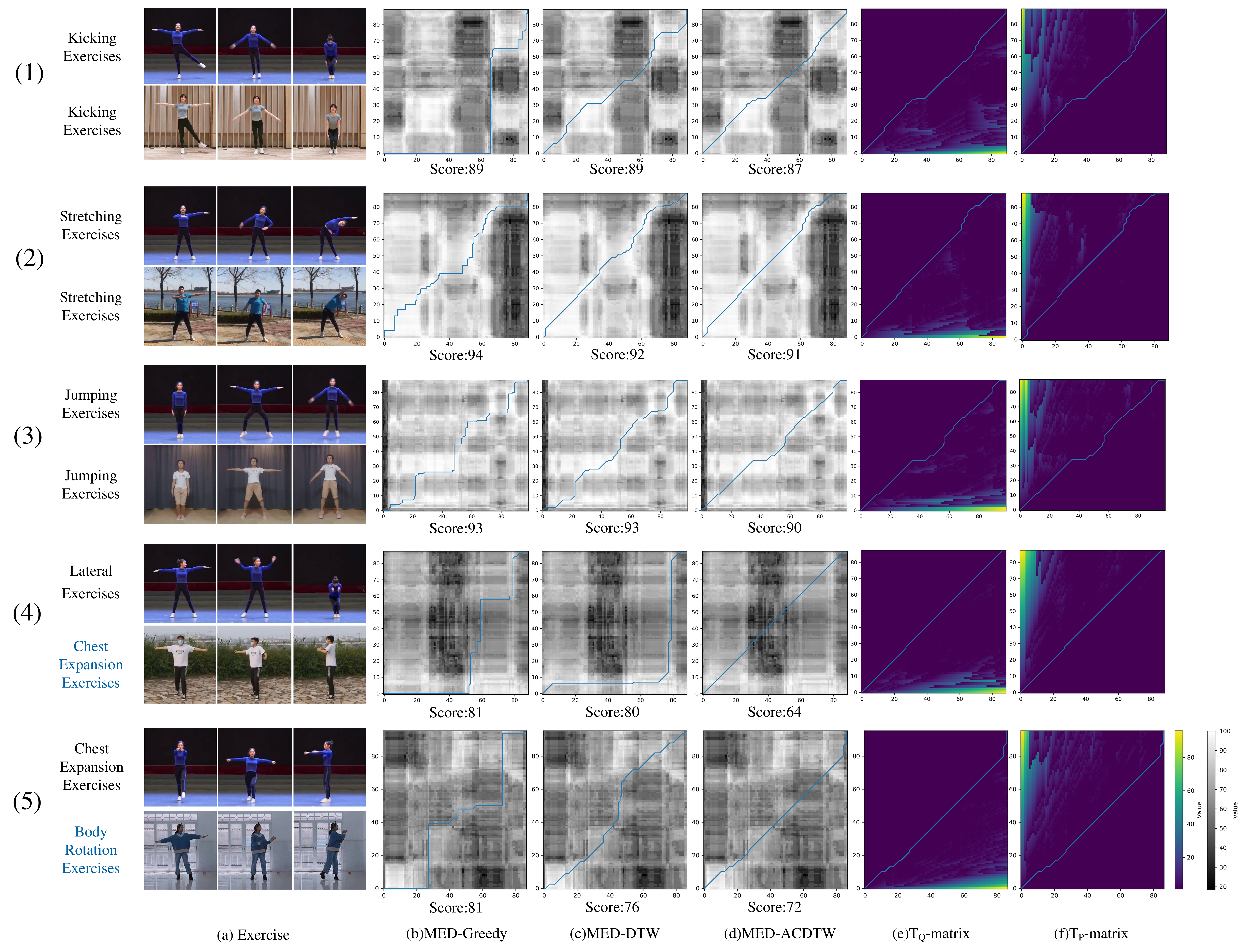}
	\caption{Action matching and score matrices. In rows (1), (2), and (3), the template videos and test videos are matched with the same action. In rows (4) and (5), the template videos and test videos are matched with different actions. Column (a) represents the exercises performed by the individual. Column (b) represents the score path obtained by the MED-Greedy method. Column (c) represents the score path obtained by the MED-DTW method. Column (d) represents the score path obtained by the MED-ACDTW method. Column (e) represents the path in the MED-ACDTW method with the $T_Q$ matrix as the background. Column (f) represents the path in the MED-ACDTW method with the $T_P$ matrix as the background.
	}
	
	\label{figure::matric}
\end{figure*}

\subsection{Parameter Ablation Experiment of MED module}
We conducted ablation experiments to examine the benefits of different algorithms, as illustrated in Table \ref{table::param_t}. In these experiments, we maintained consistent conditions by utilizing seven template videos for template samples, 153 evaluation videos for test samples, and 3D features for the feature number. All schemes employed DTW with MED.

Initially, we evaluated the influence of the hyperparameter $t$ in the MED module (as discussed in Section \ref{MED_Module}) on the experimental outcomes (see Table \ref{table::param_t}). 
The best overall performance was achieved when $t$ was set to $0.1$, resulting in Accuracy($Acc$) and $Rate80$ values of $97.39\%$ and $95.80\%$, respectively. The highest $Acc$ value of $98.04\%$ was obtained when $t$ was $0.2$. The MED module aids the algorithm in emphasizing the similarities and differences between pose features, thereby effectively evaluating motion quality.

\begin{table}[!htbp]
	\caption{Ablation study of the parameter $t$}
	\label{table::param_t}
	\centering
	\begin{tabular}{cccc}
		\toprule
		Method  & t    & Acc/\%         & Rate80/\%      \\ 
		\midrule
		\multicolumn{1}{c}{\multirow{4}{*}{MED-DTW}} & 0.1  & 97.39          & \textbf{95.80} \\
		& 0.15 & 96.73          & 92.90          \\
		& 0.2  & \textbf{98.04} & 65.00          \\
		& 0.25 & 97.39          & 42.86          \\ 
		\bottomrule
	\end{tabular}
\end{table}

\begin{table}[!htbp]
	\caption{Comparison of different features}
	\label{table::Comparison_fea}
	\centering
	\begin{tabular}{cccc}
		\toprule
		Method         & Feature        & Acc/\%         & Rate80/\%      \\ \midrule
		ED-ACDTW         & 3D        & 96.73      & 16.99      \\ \midrule
		\multicolumn{1}{c}{\multirow{4}{*}{MED-ACDTW}} 	   & 2D             & 97.39          & 82.17 \\ 
		& 3D             & 96.73          & 93.27          \\
		& 2D+ Partial 3D & 98.69 & 95.24          \\
		& 2D+3D          & \textbf{99.35}          & \textbf{95.33}          \\ \bottomrule
	\end{tabular}
\end{table}

\subsection{Ablation Experiment of the MED Module and Multi-feature Module}
As shown in Table \ref{table::Comparison_fea}, the ablation experiment initially compares the effect of the MED module with the ED (Euclidean Distance) module. The comparison reveals that while the accuracy ($Acc$) remains consistent between the two, the $Rate80$ metric significantly improves with the MED module. The high discriminability of the MED module indicates its capability to accurately distinguish variations among students in different aspects of sports skills and abilities. In different dimensions, when compared to solely using 2D features, the $Acc$ of 3D features drops by 0.7\%, but the $Rate80$ enhances by approximately 11\%. 

In terms of dimensional analysis, when 3D features are used instead of solely 2D features, there is a slight decrease in $Acc$ by 0.7\%, but a notable increase in $Rate80$ by approximately 11\%. For the integration of 2D and 3D features, the process begins with 2D centroid, angle, and orientation features, followed by the addition of 3D orientation and upper and lower limb coordination features. This combination shows that, compared to using only 3D features, the $Acc$ improves by around 1\%, and the $Rate80$ enhances by nearly 2\%.
The experiment yields the best results when combining 2D features with all available 3D features (e.g., computing angle features in both 2D and 3D). This fusion leads to an $Acc$ of 99.35\% and a $Rate80$ of 95.33\%.
This demonstrates that multi-feature fusion captures different attributes of moving characters from multiple perspectives, providing richer and more comprehensive information.

\subsection{Comparison of different methods}
Table \ref{table::Comparison_meth} showcases the experimental findings, highlighting the substantial hurdles encountered in assessing action quality within the dataset employed in this research. To ensure a level playing field, we have re-implemented these approaches grounded on the proposed MED framework. Since MED are different from their original distance modules, we add a prefix of ”MED-” to these newly implemented schemes. When all features are set to 3D, ACDTW exhibits a significant improvement compared to the MED-DTW~\cite{Ying2021openpose} and MED-Greedy~\cite{effenberg2016movement} algorithms.

In Table \ref{table::Comparison_meth}, although the $Acc$ is similar, the $Rate80$ for MED-ACDTW increases by approximately 28\% over MED-DTW. This improvement suggests that MED-ACDTW's adaptive penalty mechanisms are more effective in handling many-to-one and one-to-many mapping anomalies, thereby refining the MED-DTW path. Consequently, this enhancement allows for more distinct motion quality assessments, meeting the requirements for usability in evaluation results. In Figure \ref{figure::matric}, we visualize the experimental results. It is evident that the MED-ACDTW approach assigns lower scores to different actions, which aligns with the requirements of sports grading and enhances discriminability.  Moreover, due to the adaptive constraints, as shown in Figures \ref{figure::matric}(e) and \ref{figure::matric}(f), the curves are better suited for grading sports exercises in video form, reducing the impact of multiple frames corresponding to a single frame.

\begin{table}[!htbp]
	\caption{Comparison of action methods}
	\label{table::Comparison_meth}
	\centering
	\begin{tabular}{cccc}
		\toprule
		Method     & Feature & Acc/\%         & Rate80/\%      \\ \midrule
		MED-DTW~\cite{Ying2021openpose}    & 3D      & 98.04          & 65.00 \\          
		\midrule
		MED-Greedy~\cite{effenberg2016movement} & 3D      & 94.12          & 79.93          \\
		\midrule
		\multicolumn{1}{c}{\multirow{2}{*}{MED-ACDTW}}  & 3D      & 96.73 & 93.27          \\
		& 2D+3D   & \textbf{99.35}          & \textbf{95.33}          \\ \bottomrule
	\end{tabular}
\end{table}

In addition, this work includes a comparison with the currently popular regression-based approaches, as shown in Table \ref{table::Comparison_regression}. It employs two different data-splitting strategies. The first involves randomly selecting 22 videos from the dataset as the test set. The second strategy entails extracting three 3-second videos from a single action, using these videos as the test set, and ensuring the training set does not include the corresponding actions for this specific character's movements. This results in 21 videos comprising the test set. In Table \ref{table::Comparison_meth}, we denote the second partitioning strategy with an asterisk (*).

\begin{figure}[htbp]
	\centering
	\includegraphics[width=240pt]{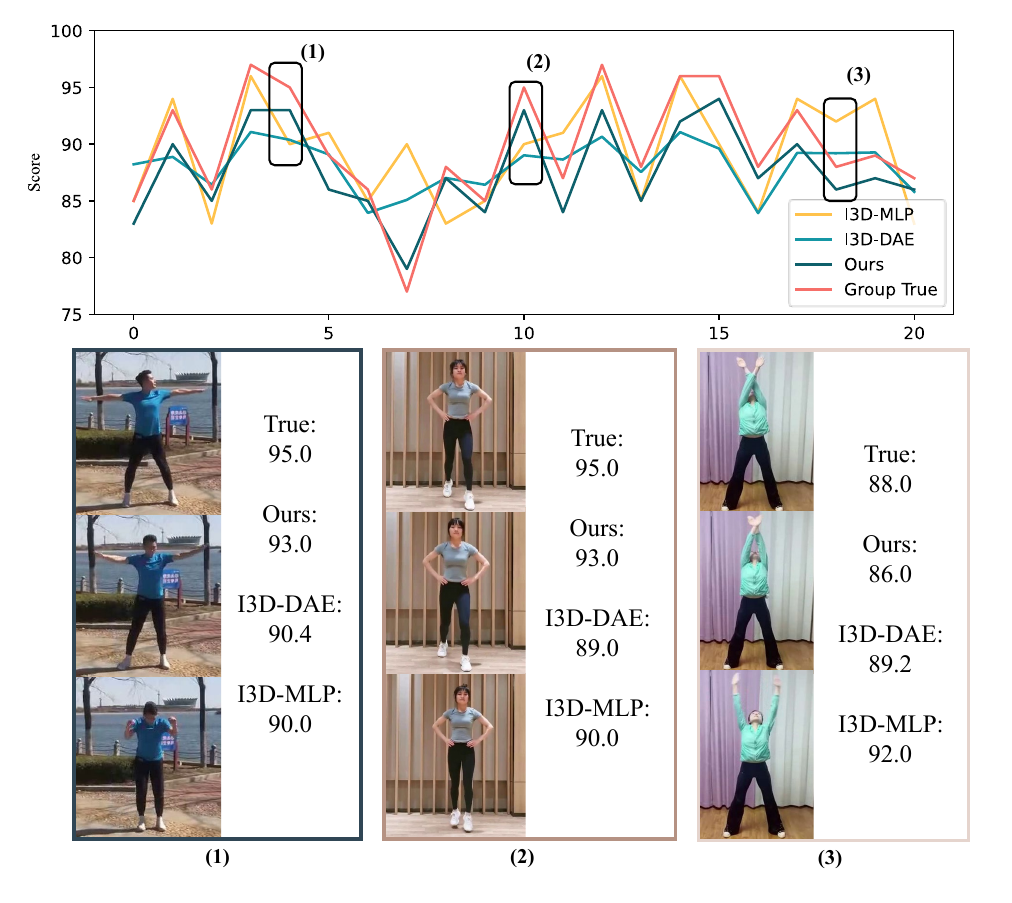}
	\caption{Comparison of popular methods. (1) represents lateral exercises. (2) represents jumping exercises. (3) represents abdominal and back exercises.}
	\label{figure::popular}
\end{figure}

In the case where there are similar videos in the training set, the Spearman correlation of MED-ACDTW improved by approximately 2\% compared to the popular I3D-MLP method. It exhibited an improvement of about 0.4\% against I3D-DAE. Compared to the HGCN method, there was an increase of around 4\%. In the second strategy, the approach demonstrates significant enhancement over the existing methods, with an improvement of approximately 28\% compared to I3D-MLP, and an improvement of approximately 10\%. Compared to the HGCN method, there was an increase of around 12\%. This indicates that MED-ACDTW is capable of obtaining accurate scores even when encountering untrained actions, as long as there are template videos available. In contrast, other regression methods show poorer perceptual ability for actions that have not been trained on. Figure \ref{figure::popular} illustrates the performance of the methods mentioned in Table \ref{table::Comparison_regression} in terms of evaluated scores on the test set. Additionally, three specific points are selected for visualization. The curve of the proposed approach exhibits fluctuations that are closer to the actual data.

\begin{table}[!htbp]
	\caption{Comparison of regression methods}
	\label{table::Comparison_regression}
	\centering
	\begin{tabular}{ccc}
		\toprule
		Method  & Sp. Corr.      & Sp. Corr.$^{*}$     \\ \midrule
		I3D-MLP~\cite{tang2020uncertainty} & 0.9273          & 0.6699          \\
		I3D-DAE\cite{zhang2023auto} & 0.9376          & 0.8465          \\
		HGCN\cite{zhou2023hierarchical} & 0.9064          & 0.8191          \\
		MED-ACDTW    & \textbf{0.9418} & \textbf{0.9427} \\ \bottomrule
	\end{tabular}
\end{table}

\subsection{Analysis of action quality assessment}
\begin{figure}[ht]
	\centering
	\includegraphics[width=240pt]{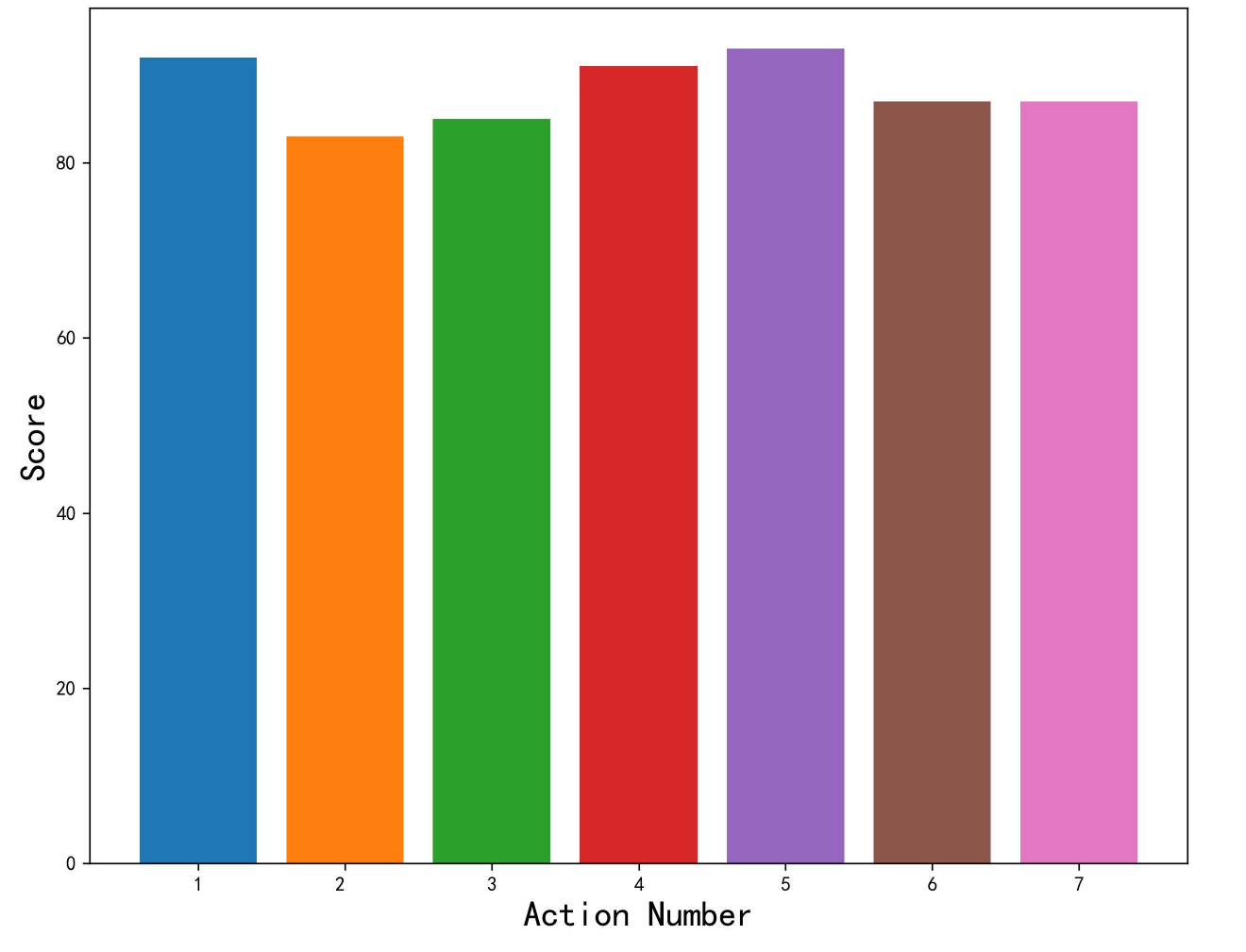}
	\caption{Action score analysis of each section}
	\label{figure::res7}
\end{figure}

\begin{figure}[ht]
	\centering
	\includegraphics[width=240pt]{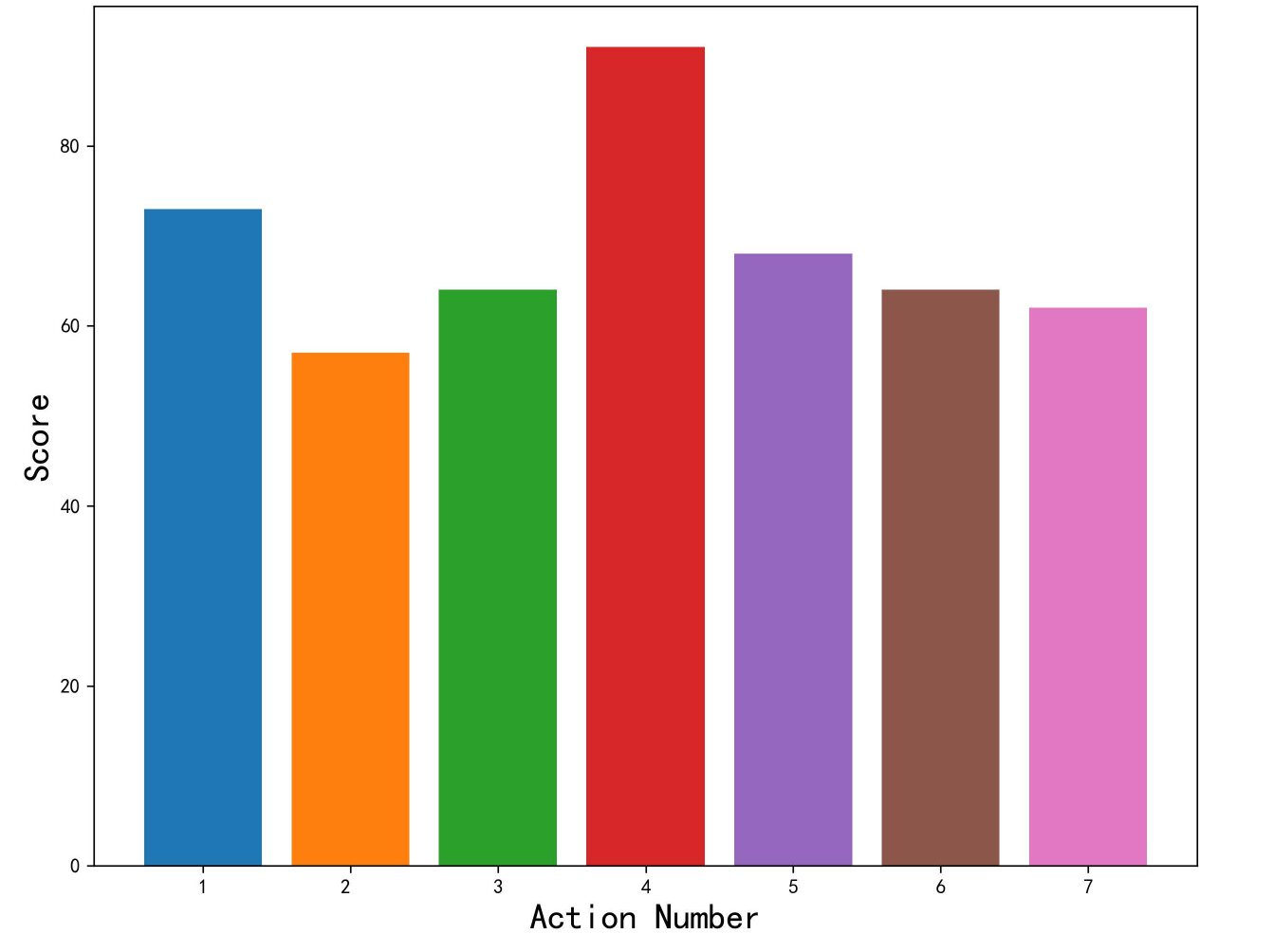}
	\caption{Score analysis of stretching exercises versus other exercises}
	\label{figure::res8}
\end{figure}

Qualitative results from the broadcast calisthenics dataset using 2D+3D feature for MED-ACDTW are shown in Figure \ref{figure::res7} and \ref{figure::res8}. Our visualization scheme uses a hybrid feature for MED-ACDTW, with action names and serial numbers as shown in Table \ref{table::dataset}. Since each exercise in the dataset takes three segments of about 3 seconds, we use the average of the three segments as the score obtained by the test subject in that exercise. As can be seen from the Figure \ref{figure::res7}, test subjects who take the exercises seriously have scores above 80 for each exercise, and the best matching performances are for actions with serial numbers 1 and 5.

Then, we analyze the scoring of stretching motion templates against different motions in Figure \ref{figure::res8}. Each motion corresponds to a score obtained from the stretching motion. 
The highest score is indeed for the stretching motion in number 5, while others score below 80.
The chest-expanding motion in number 2, which is quite different, scores below 60, while other more similar motions score around 75. This proves the rationality of the differentiation in motion quality assessment, making this algorithm effectively solve the practical problems of motion posture assessment in our study.

\section{Conclusion}
This study introduces a novel action quality assessment algorithm called MED-ACDTW. In the aspect of feature extraction, we find that joint angles, barycentric, directions, and upper-lower limb linkage features significantly contribute to our scoring system. These four feature categories are obtained by transforming 2D and 3D coordinates using the relationships between each node. Furthermore, unlabeled approaches do not heavily depend on dataset labeling.
In the DTW algorithm, we have made modifications to the path score and incorporated adaptive constraints, which markedly enhance our discriminatory capabilities and mitigate the impact of one-to-many and many-to-one occurrences in the action time series. Essentially, our MED-ACDTW scheme effectively conducts action quality assessments. In the future, we aim to refine existing features or introduce additional ones to more effectively distinguish between similar movements.

Additionally, we anticipate enriching our dataset with more complex data to expand our gymnastics dataset and extend our model to cover other sports categories. These represent promising directions for future research.

\end{document}